\documentclass{article}
\pdfoutput=1
\usepackage{cancel}
\usepackage{spconf}
\usepackage{enumerate,color,multirow,graphicx,booktabs,array,stmaryrd,rotating,hyperref}
\usepackage{amsbsy, amstext, amssymb, amsthm, amsmath, bbm, bm}
\usepackage{algorithm,algorithmic}

\newcommand{\diag}{\mathrm{diag}}

\newcommand{\Abf}{{\bm A}}

\newcommand{\Hbf}{{\bm H}}

\newcommand{\Wbf}{{\bm W}}
\newcommand{\Xbf}{{\bm X}}

\newcommand{\xbf}{{\bm x}}

\newcommand{\zbf}{{\bm z}}

\title{Generative Graph Convolutional Network for Growing Graphs}
\name{Da Xu, Chuanwei Ruan, Kamiya Motwani, Evren Korpeoglu, Sushant Kumar, Kannan Achan}
\address{Walmart Labs, Sunnyvale, California, USA}

\begin{document}
\maketitle
\begin{abstract}
Modeling generative process of growing graphs has wide applications in social networks and recommendation systems, where cold start problem leads to new nodes isolated from existing graph. Despite the emerging literature in learning graph representation and graph generation, most of them can not handle isolated new nodes without nontrivial modifications. The challenge arises due to the fact that learning to generate representations for nodes in observed graph relies heavily on topological features, whereas for new nodes only node attributes are available. Here we propose a unified generative graph convolutional network that learns node representations for all nodes adaptively in a generative model framework, by sampling graph generation sequences constructed from observed graph data. We optimize over a variational lower bound that consists of a graph reconstruction term and an adaptive Kullback-Leibler divergence regularization term. We demonstrate the superior performance of our approach on several benchmark citation network datasets.
\end{abstract}


\section{Introduction}
\label{sec:intro}
\subsection{Background}
\label{sec:background}

Real-world graph structured data is often under dynamic growth as new nodes emerge over time. However, directly modeling the generation process from observed graph data remains a difficult task due to the complexity of graph distributions. Recently, there have been significant advances in learning graph representations by mapping nodes onto latent vector space. The latent factors (embeddings) which have simpler geometric structures can then be used for downstream machine learning analysis such as generating graph structures \cite{li2018learning} and various semi-supervised learning tasks \cite{yang2016revisiting}.

Some early approaches in node embedding such as \textsl{graph factorization algorithm} \cite{ahmed2013distributed}, \textsl{Laplacian eigenmaps} \cite{belkin2002laplacian} and \textsl{HOPE} \cite{ou2016asymmetric} are based on deterministic matrix-factorization techniques. Later approaches arise from random walk techniques that provide stochastic measures for analysis, including \textsl{DeepWalk} \cite{perozzi2014deepwalk}, \textsl{node2vec} \cite{grover2016node2vec} and \textsl{LINE} \cite{tang2015line}. More recent graph embedding techniques focus on building deep graph covolutional networks (\textsl{GCN}) \cite{kipf2016semi} as encoders that aggregate neighborhood information \cite{scarselli2009graph}. Variants of \textsl{GCN} have been proposed to tackle the computational complexity for large graphs, such as \textsl{FastGCN} \cite{chen2018fastgcn} which applies graph sampling techniques. \textsl{GraphSAGE} \cite{hamilton2017inductive} is another time-efficient inductive graph representation learning approach that implements localized neighborhood aggregations.    

On the other side, advancements in generative models compatible with deep neural networks such as variational autoencoders (\textsl{VAE}) \cite{kingma2013auto,rezende2014stochastic} and generative adversarial networks (\textsl{GAN}) \cite{goodfellow2014generative} have enabled direct modeling for generation of complex distributions. As a consequence, there have been several recent work on deep generative models for graphs \cite{you2018graphrnn,simonovsky2018graphvae,kipf2016variational,grover2018graphite}. However, many of them only deal with fixed graphs \cite{grover2018graphite, kipf2016variational} or graphs of very small sizes \cite{simonovsky2018graphvae,li2018learning}. Moreover, most graph representation learning methods require at least some topological features from all nodes in order to conduct neighborhood aggregations or random walks, which is clearly infeasible for growing graphs with isolated new nodes. To obtain embeddings and further generate graph structures for both new and old nodes, it is essential to utilize node attributes. Also, instead of learning how the observed graph is generated as a whole, the graph generation should be decomposed into sequences that reflect how new nodes are sequentially fitted into existing graph structures.

\subsection{Related methods}
\label{sec:related-methods}

\textbf{Variational Autoencoder} Unlike vanilla autoencoder, VAE treats the latent factors $\zbf$ as random variables such that they can capture variations in the observed data $\xbf$ \cite{kingma2013auto}. \textsl{VAE} has shown high efficiency in recovering complex multimodal data distributions. The parameters in encoding distribution $q_{\phi}(\zbf | \xbf)$ and decoding distribution $p_{\theta}(\xbf | \zbf)$ are optimized over the evidence (variational) lower bound (ELBO)
$$
\log p(\xbf) \geq E_{q_{\phi}(\zbf | \xbf)}[\log p_{\theta}(\xbf|\zbf)] - KL(q_{\phi}(\zbf | \xbf) || p_0(\zbf)).
$$
The expectation with respect to $q_{\phi}(\zbf | \xbf)$ is approximated stochastically by reparametrizing $\zbf$ as $\bm \mu + \bm \sigma \odot \bm \epsilon$, where $\bm \epsilon$ are independent standard Gaussian variables. This is also referred to as 'reparameterization trick' \cite{kingma2013auto}. It allows sampling directly from $\zbf$ so that the backpropagation technique becomes feasible for training deep networks. 

\textbf{Graph Convolutional Network} The original \textsl{GCN} \cite{kipf2016semi} deals with node classification as a semi-supervised learning task. The layer-wise propagation rule is defined as $\Hbf^{l+1} = \sigma(\hat{\Abf}\Hbf^{l}\Wbf_l)$. Here $\hat{\Abf}$ is the normalized adjacency matrix with $\hat{\Abf}_{i,j} = \frac{\Abf_{i,j}}{\sqrt{deg(i)deg(j)}}$ where $deg(i)$ gives the degree of node $i$. The $\sigma(.)$ is some activation function such as ReLU. $\Hbf^l$ is the output of the $(l-1)^{th}$ layer and $\Wbf_l$ is the layer-specific aggregation weights. Here $\Wbf_l \in \mathbf{R}^{d_{l} \times d_{l+1}}$ where $d_l$ is the dimension of the hidden units on $l^{th}$ layer.

\textbf{Graph Convolutional Autoencoder (\textsl{GAE})} \textsl{GAE} is an important extension of \textsl{GCN} for learning node representations for link prediction \cite{kipf2016variational}. A two-layer \textsl{GCN} is used as encoder 
$\zbf = GCN(\Abf, \Xbf) = \hat{\Abf}ReLu(\hat{\Abf}\Xbf \Wbf_0)\Wbf_1$.
When adapting \textsl{GCN} into \textsl{VAE} framework (\textsl{GCN-VAE}), the hidden factors $\zbf$ are assumed to follow independent normal distributions which are parameterized by mean $\bm \mu$ and log of standard deviation $\bm \sigma$, where $\bm \mu = GCN_{\mu}(\Abf, \Xbf)$ and $\bm \sigma = GCN_{\sigma}(\Abf, \Xbf)$. The pairwise decoding (generative) distribution for link between node $i$ and $j$ is simply $f(\langle \zbf^i, \zbf^j \rangle)$ where $f(.)$ is the sigmoid function. The ELBO is formulated as $E_{q(\zbf|\Xbf, \Abf)}[\log p(\Abf | \zbf)] - KL(q(\zbf | \Abf, \Xbf) || p_0(\zbf)).$ 

\textbf{\textsl{GraphRNN}}
Recently a graph generation approach relying only on topological structure was proposed in \cite{you2018graphrnn}. It learns the sequential generation mechanism by training on a set of sampled sequences of decomposed graph generation process. The formation of each new edge is conditioned on the graph structure generated so far. Our approach refers to this idea of sampling from decomposed generation sequences.

\subsection{Present Work}
\label{sec:present-work}

This work addresses the challenge of generating graph structure for growing graphs with new nodes that are unconnected to the previous observed graph. It has important meaning for the cold start problems \cite{lika2014facing} in social networks and recommender systems. The major assumption is that the underlying generating mechanism is stationary during growth. \textsl{GraphRNN} neither takes advantage of node attributes nor does it naturally extends to isolated new nodes. Most other graph representation learning methods have similar issues, specifically the isolation from existing graph hinders passing messages or implementing aggregation.

We deal with this problem by learning how graph structures are generated sequentially, for cases where both node attributes and topological information exist as well as for cases where only node attributes are available. To the best of our knowledge, this work is the first of its kind in graph signal processing.

\section{Method}
\label{sec:method}
Let the input be observed undirected graph $\mathcal{G}=(\mathcal{V},\mathcal{E})$ with associated binary adjacency matrix $\Abf$, node attributes $\Xbf \in \mathbb{R}^{n \times d_0}$ and the new nodes $\mathcal{V}^{new}$ with attributes $\Xbf^{new}$. Our approach learns the generation of overall adjacency matrix $\Abf^{new}$ for $\mathcal{V} \cup \mathcal{V}^{new}$. 

\subsection{Proposed Approach}
\label{sec:proposed-approach}

We start by treating incoming nodes as being added one-by-one into the graph. Let $\Abf^{\pi}_{t} \in \mathbb{R}^{t\times t}$ be the observed adjacency matrix up to the $t^{th}$ step according to the ordering $\pi$. When the $(t+1)^{th}$ node is presented, we treat it as connected to all of previous nodes with the same probability $\tilde{p}$, where $\tilde{p}$ may reflect the overall sparsity of the graph. Hence the new candidate adjacency matrix denoted by $\tilde{\Abf}^{\pi}_{t+1}$ is given by
\begin{equation}
\label{eqn:adj-prior}
\begin{split}
&(\tilde{\Abf}^{\pi}_{t+1})_{t+1,t+1} = 1,  (\tilde{\Abf}^{\pi}_{t+1})_{1:t,1:t} = \Abf^{\pi}_{t},\\
&p((\tilde{\Abf}^{\pi}_{t+1})_{k,t+1} = 1) = \tilde{p} \text{ for } k = 1,2,\cdots, t.
\end{split}
\end{equation}

Similar to \textsl{GraphRNN}, we obtain the marginal distribution for graph by sampling the auxiliary $\pi$ from the joint distribution of $p(\mathcal{G}, (\Abf^{\pi}, \Xbf^{\pi}))$ with
$$p(\mathcal{G}) = \sum_{\pi}p((\Abf^{\pi}, \Xbf^{\pi})\mathbbm{1}[f_G(\Abf^{\pi}, \Xbf^{\pi}) = \mathcal{G}]),$$
where $f_{G}(\Abf^{\pi}, \Xbf^{\pi})$ maps the tuple $(\Abf^{\pi}, \Xbf^{\pi})$ back to a unique graph $\mathcal{G}$. 
Each sampled $\pi$ gives a $(\Abf^{\pi}, \Xbf^{\pi})$ that constitutes one-sample mini-batch that drives the stochastic gradient descent (SGD) for updating parameters.

To illustrate the sequential generation process, we decompose joint marginal log-likelihood of $(\Abf_{\leq n}, \Xbf_{\leq n})$ under the node ordering $\pi$ into
\begin{equation}
\begin{split}
\log p(\Abf^{\pi}_{\leq n},\Xbf^{\pi}_{\leq n}) &= \sum_{i=1}^{n-1} \log p(\Abf^{\pi}_{\leq i+1},\Xbf^{\pi}_{\leq i+1} | \Abf^{\pi}_{\leq i},\Xbf^{\pi}_{\leq i}) \\
& + \log p(\Abf^{\pi}_1, \Xbf^{\pi}_1).
\end{split}
\end{equation}
The log-likelihood term of initial state $\log p(\Abf^{\pi}_1, \Xbf^{\pi}_1)$ is not of interest as we focus on modeling transition steps.

Following \textsl{VAE} framework with hidden factors as Gaussian variables, for each transition step, we use encoding distribution $q^i_{\phi}(\zbf|\Abf_{\leq i}, \Xbf_{\leq i})$, generating distribution $p^i_{\theta}(\Abf| \zbf^i)$, and conditional prior $p^i_0(\zbf | \Abf_{\leq i}, \Xbf_{\leq i})$. From now on we treat the conditional on $\pi$ as implicit for simplicity of notation. The variational lower bound for each step is given by:
\begin{equation}
\label{eqn:ELBO}
\begin{split}
&\log p(\Abf_{\leq i+1}, \Xbf_{\leq i+1} | 
\Abf_{\leq i}, \Xbf_{\leq i}) \\
& \geq E_{q^i_{\phi}(\zbf^{i+1} | \tilde{\Abf}_{\leq i+1}, \Xbf_{\leq i+1})} [\log p^i_{\theta}(\Abf_{\leq i} | \zbf^i)] \\
& - KL(q^i_{\phi}(\zbf^{i+1} | \tilde{\Abf}_{\leq i+1}, \Xbf_{\leq i+1}) \| p_0^{i}(\zbf^{i+1} | \Abf_{\leq i}, \Xbf_{\leq i})) + C \\
& \equiv ELBO_i + C.
\end{split}
\end{equation}
Here $C=\log \int_{\mathcal{Z}} p^i_{\theta}(\Xbf_{\leq i} | \zbf^i) q^i_{\phi}(\zbf^{i+1} | \tilde{\Abf}_{\leq i+1}, \Xbf_{\leq i+1}) d\nu(\zbf)$ is the reconstruction term for node attributes, which is not our target. We will discuss the interpretation for our evidence lower bound in Section \ref{sec:sequantial-elbo}. Given that we have assumed the consistency of underlying generating mechanism, we use the same set of parameters for each step.

When formulating encoding distribution, due to the efficiency of \textsl{GCN} in node classification and linkage prediction, we adopt their convolutional layers. The two-layer encoder for the $i^{th}$ step is then given by:
\begin{equation}
\label{eqn:encoding-layer}
\begin{split}
&\bm \mu(\zbf^i | \Xbf_{\leq i+1}) = \hat{\tilde{\Abf}}_{\leq i+1} \sigma(\hat{\tilde{\Abf}}_{\leq i+1} \Xbf_{\leq i+1} \Wbf_{0})\Wbf_1, \\
& \diag (\bm \Sigma(\zbf^i | \Xbf_{\leq i+1}) = \hat{\tilde{\Abf}}_{\leq i+1} \sigma(\hat{\tilde{\Abf}}_{\leq i+1} \Xbf_{\leq i+1} \Wbf_{0})\Wbf_2, 
\end{split}
\end{equation}
where $\sigma(.)$ is activation function and $\hat{\tilde{\Abf}}$ denotes the normalized candidate adjacency matrix constructed by (\ref{eqn:adj-prior}).
We also adopt the pairwise inner product decoder for edge generation:
\begin{equation}
\label{eqn:decoding-layer}
p_{i,j} = p(\Abf_{i,j} = 1 | \zbf_i, \zbf_j) = f(\langle \zbf_i, \zbf_j \rangle),
\end{equation}
with $f(.)$ being the sigmoid function. Another reason for using simple decoder being that in \textsl{VAE} framework if the generative distribution is too expressive, the latent factors are often ignored \cite{chen2016variational}. 

As for conditional priors of hidden factors, standard Gaussian priors are no longer suitable because we already have information from previous $i$ nodes at the $(i+1)^{th}$ step. Hence, we use what the model has informed us till the $i^{th}$ step in an adaptive way by treating $\zbf^{i+1} \in \mathbb{R}^{(i+1) \times d_2}$ as $[\zbf^{i+1}_{1:i}, \zbf^{i+1}_{i+1}]$, where $\zbf^{i+1}_{1:i}$ are the hidden factors for previous nodes and $\zbf^{i+1}_{i+1}$ is for the new node. For $\zbf^{i+1}_{1:i}$ we can use the encoding distribution $p_{\phi}(\zbf_{1:i} | \tilde{\Abf}_{\leq i+1}, \Xbf_{\leq i+1})$ where the candidate adjacency matrix $\tilde{\Abf}_{\leq i}$ passes information from previous steps. For the new node we keep using standard Gaussian prior. This gives us
\begin{equation}
\label{eqn:hidden-prior}
\begin{split}
&p_0^i(\zbf^{i+1}_{1:i} | \Abf_{\leq i}, \Xbf_{\leq i}) = p_{\phi}(\zbf_{1:i} | \tilde{\Abf}_{\leq i+1}, \Xbf_{\leq i+1}) \\
&\zbf^{i+1}_{i+1} | \Abf_{\leq i}, \Xbf_{\leq i} \sim N(0,I)
\end{split}
\end{equation}
We use the sum of negative ELBO in each transition step as loss function ($L = -\sum_{i=1}^{n-1}ELBO_i$) and obtain optimal aggregation weights $[\Wbf_0, \Wbf_1, \Wbf_2]$ by minimizing this loss.

In practice, it's not necessary to consider adding new nodes one-by-one. Instead, the new nodes can be added in a batch-wise fashion to alleviate computational costs. In preliminary experiments we also observe that sampling uniformly at random from all node permutations gives very similar results to sampling from BFS orderings, hence we report results with the uniform sampling schema. 


\begin{figure*}[h!]
\label{fig:workflow}
  \caption{An illustration for the workflow of our approach at a transition step. The growing graph has an observed subgraph with three nodes (node 0, 1 and 2) and an incoming new node (node 3). The informative conditional priors for latent factors $\zbf^{(3)}_{0:2}$ contain structural information of the observed subgraph. The encoding distributions for all latent factors are formed according to the candidate adjacency matrix where candidate edges (the dashed edges) are added to original subgraph.}
  \centering
  \includegraphics[width=0.9\textwidth]{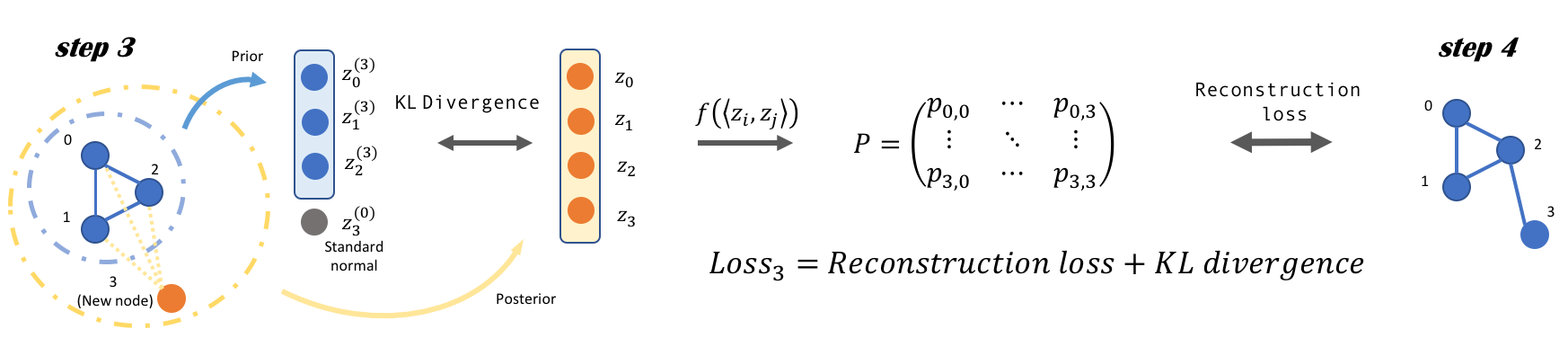}
\end{figure*}

\subsection{Adaptive Evidence Lower Bound}
\label{sec:sequantial-elbo}
The loss function can be rearranged into (\ref{eqn:adaptive-elbo}) (with $\beta=1$):
\begin{equation}
\label{eqn:adaptive-elbo}
\begin{split}
& -\sum_{i=1}^{n-1}E_{q^i_{\phi}(\zbf^{i+1} | \tilde{\Abf}_{\leq i+1}, \Xbf_{\leq i+1})} [\log p_{\theta}(\Abf_{\leq i} | \zbf^i) ] \\
& + \beta \sum_{i=1}^{n-1} KL(q^i_{\phi}(\zbf^{i+1} | \tilde{\Abf}_{\leq i+1}, \Xbf_{\leq i+1}) \| p_0^{i}(\zbf^{i+1} | \Abf_{\leq i}, \Xbf_{\leq i})).
\end{split}
\end{equation}
The first term sums up the reconstruction loss in each generation step. The second term serves as an adaptive regularizer that enforces the posterior of latent factors for observed nodes to remain close to their priors which contain information from previous steps. This can prevent the model from overfitting the edges of the new nodes, which is quite helpful in our batched version where new edges can outnumber original edges, as we are fitting new nodes into original structure.

Similar to $\beta$-\textsl{VAE} \cite{higgins2017beta}, we also introduce the tuning parameter $\beta$ as shown in (\ref{eqn:adaptive-elbo}) to control the tradeoff between the reconstruction term and the adaptive regularization.

\section{Experiment}
\label{sec:experiment}
\begin{table*}[btp]
\footnotesize	
\centering
\caption{Results for link prediction tasks in citation networks. Standard error is computed over 10 runs with random initializations on random dataset splits. The first three rows are results for the first task on new nodes, and last three rows are results for the second task on nodes in observed graph.}
\begin{tabular}{c c c c c c c}
\hline
\multirow{ 2}{*}{\textbf{Method}} & \multicolumn{2}{c}{\textbf{Cora}} & \multicolumn{2}{c}{\textbf{Citeseer}} & \multicolumn{2}{c}{\textbf{Pubmed}} \\
& AUC & AP  & AUC & AP & AUC & AP \\ \hline 
\multicolumn{7}{c}{\textbf{Isolated new nodes}} \\ 
GCN-VAE & 75.12 $\pm$ 0.4 & 76.32 $\pm$ 0.3 & 79.36 $\pm$ 0.3 & 82.13 $\pm$ 0.1  & 85.52 $\pm$ 0.2 & 85.43 $\pm$ 0.1\\ 
MLP-VAE & 75.59 $\pm$ 0.7 & 75.64 $\pm$ 0.5 & 81.76 $\pm$ 0.6 & 83.67 $\pm$ 0.4 & 77.13 $\pm$ 0.4 & 77.24 $\pm$ 0.3 \\
G-GCN & \textbf{83.30} $\pm$ 0.3 &\textbf{85.03} $\pm$ 0.3 & \textbf{89.54} $\pm$ 0.2 & \textbf{91.30} $\pm$ 0.2  & \textbf{87.49} $\pm$ 0.2 & \textbf{87.24} $\pm$ 0.1 \\ \hline
\multicolumn{7}{c}{\textbf{Nodes in observed graph}} \\ 
GCN-VAE & 93.15 $\pm$ 0.4 & 94.42 $\pm$ 0.2 &  93.27 $\pm$ 0.4 & 94.42 $\pm$ 0.1  & 96.74 $\pm$ 0.4 & 96.94 $\pm$ 0.3\\ 
MLP-VAE & 86.55 $\pm$ 0.2 & 87.21 $\pm$ 0.3 & 87.13 $\pm$ 0.2 & 89.34 $\pm$ 0.1 &  79.39 $\pm$ 0.5 &  79.53 $\pm$ 0.3 \\
G-GCN & \textbf{94.07} $\pm$ 0.4 & \textbf{95.15} $\pm$ 0.2 & \textbf{94.62} $\pm$ 0.7 & \textbf{95.93} $\pm$ 0.7  & \textbf{96.96} $\pm$ 0.6 & \textbf{97.27} $\pm$ 0.5 \\ \hline
\end{tabular}
\label{tab:new_nodes}
\end{table*}

We test our generative graph convolution network (\textsl{G-GCN}) for growing graphs on two tasks: link prediction for isolated new nodes, and for nodes in observed graph. We use three benchmark citation network datasets: Cora, Citeseer and Pubmed. Their details are described in \cite{sen2008collective}. Node attributes are informative for all three datasets, which is indicated by the results of \textsl{GCN-VAE} in \cite{kipf2016variational}.

\subsection{Baselines}
\label{sec:baselines}

We compare our approach against \textsl{GCN-VAE} and a multilayer perceptron \textsl{VAE} (\textsl{MLP-VAE}) \cite{kingma2013auto}. Here the encoder of \textsl{MLP-VAE} is constructed by replacing the adjacency matrices in \textsl{GCN-VAE} with non-informative identity matrices. Their decoders are the same as our approach in (\ref{eqn:decoding-layer}). The difference is that \textsl{GCN-VAE} uses both topological information and node attributes, while \textsl{MLP-VAE} only considers node attributes. When predicting edges for isolated new nodes, for all three methods, we plug the 'candidate' adjacency matrix $\tilde{\Abf}$ formulated in (\ref{eqn:adj-prior}) with $\tilde{p}=0$ into the encoder-decoder frameworks and recover the true adjacency matrix. 

We choose these two methods to compare with, instead of others, because both of them are able to utilize node attributes and follow from \textsl{VAE} framework. As we mentioned, most other graph embedding and graph generation techniques do not work for growing graphs without nontrivial modifications.

\subsection{Experiment Setup}
\label{sec:experiment-setup}

\textbf{Link prediction for isolated new nodes} \\
For each citation network, a growing graph is constructed by randomly sampling an observed subgraph containing 70\% of all nodes. The left-out nodes are treated as isolated new nodes. The subgraph is used for training and the validation and test sets are formed by the edges between nodes in observed subgraph and the new nodes as well as the edges among the new nodes according to the original full graph. As we are treating new nodes as being added in a batch-wise fashion, the size of new node batch is set to be $\frac{\# \{\text{training nodes}\}}{3}$. \\
\textbf{Link prediction for nodes in observed graph} \\
We then test our model on the original link predictions task \cite{kipf2016variational}, which predicts the existence of unseen edges between nodes in observed graph.  
In this task we adopt their experiment setup, where 10\% and 5\% of the edges are removed from the training graph and used as positive validation set and test set respectively. The same amount of unconnected node pairs are sampled and constitute the negative examples. 

In both tasks we use a 400-dim hidden layer and 200-dim latent variables, and train for 200 iterations using the \textsl{Adam} optimizer with a learning rate of 0.001 for all methods. Notice that all three methods use encoding layers of the same form and their decoding layers are all parameter-free, so they already have the same number of parameters. The implementation of \textsl{GCN-VAE} on the second task is conducted using their official Tensor-Flow code. The rest are conducted with our own  implementations with PyTorch.

\subsection{Results}
\label{sec:results}

We report \textsl{area under the ROC curve} (AUC) and \textsl{average precision} (AP) scores for each model on the test sets for the two tasks (Table \ref{tab:new_nodes}). 

Firstly, our approach outperforms both baselines in new node link prediction task across all three datasets, in terms of both AUC and AP. By comparing to \textsl{MLP-VAE} we show our advantage of learning with topological information, and our better performance over \textsl{GCN-VAE} indicates the importance of modeling the sequential generating process when making predictions on new nodes.


Secondly, \textsl{G-GCN} has comparable or even slightly better results than \textsl{GCN-VAE} on link prediction task in observed graph, which suggests that our superior performance on isolated new nodes is not at the cost of the performance on nodes in observed graph. This is within expectation since our approach learns the generation process as graph structure keeps growing under our sequential training setup, where new nodes are added in each step. It targets on nodes in observed graph as well as new nodes while not overfitting either of them. In a nutshell, our approach achieves better performance on link prediction task for the growing graphs as a whole. 

\section{Conclusion and Future Work}
\label{sec:conclusion}
We propose a generative graph convolution model for growing graphs that incorporates graph representation learning and graph convolutional network into a sequential generative model. Our approach outperforms others in all benchmark datasets on link prediction for growing graphs.

However, scalability remains a major issue as the computational complexity depends on the size of full graph. The idea of localized convolution from \textsl{GraphSAGE} \cite{hamilton2017inductive} and graph sampling from \textsl{FastGCN} \cite{chen2018fastgcn} may have pointed out promising directions, which we leave to future work.

\vfill\pagebreak
\bibliographystyle{IEEEbib}
\bibliography{reference}

\begin{thebibliography}{10}

\bibitem{li2018learning}
Yujia Li, Oriol Vinyals, Chris Dyer, Razvan Pascanu, and Peter Battaglia,
\newblock ``Learning deep generative models of graphs,''
\newblock {\em arXiv preprint arXiv:1803.03324}, 2018.

\bibitem{yang2016revisiting}
Zhilin Yang, William~W Cohen, and Ruslan Salakhutdinov,
\newblock ``Revisiting semi-supervised learning with graph embeddings,''
\newblock {\em arXiv preprint arXiv:1603.08861}, 2016.

\bibitem{ahmed2013distributed}
Amr Ahmed, Nino Shervashidze, Shravan Narayanamurthy, Vanja Josifovski, and
  Alexander~J Smola,
\newblock ``Distributed large-scale natural graph factorization,''
\newblock in {\em Proceedings of the 22nd international conference on World
  Wide Web}. ACM, 2013, pp. 37--48.

\bibitem{belkin2002laplacian}
Mikhail Belkin and Partha Niyogi,
\newblock ``Laplacian eigenmaps and spectral techniques for embedding and
  clustering,''
\newblock in {\em Advances in neural information processing systems}, 2002, pp.
  585--591.

\bibitem{ou2016asymmetric}
Mingdong Ou, Peng Cui, Jian Pei, Ziwei Zhang, and Wenwu Zhu,
\newblock ``Asymmetric transitivity preserving graph embedding,''
\newblock in {\em Proceedings of the 22nd ACM SIGKDD international conference
  on Knowledge discovery and data mining}. ACM, 2016, pp. 1105--1114.

\bibitem{perozzi2014deepwalk}
Bryan Perozzi, Rami Al-Rfou, and Steven Skiena,
\newblock ``Deepwalk: Online learning of social representations,''
\newblock in {\em Proceedings of the 20th ACM SIGKDD international conference
  on Knowledge discovery and data mining}. ACM, 2014, pp. 701--710.

\bibitem{grover2016node2vec}
Aditya Grover and Jure Leskovec,
\newblock ``node2vec: Scalable feature learning for networks,''
\newblock in {\em Proceedings of the 22nd ACM SIGKDD international conference
  on Knowledge discovery and data mining}. ACM, 2016, pp. 855--864.

\bibitem{tang2015line}
Jian Tang, Meng Qu, Mingzhe Wang, Ming Zhang, Jun Yan, and Qiaozhu Mei,
\newblock ``Line: Large-scale information network embedding,''
\newblock in {\em Proceedings of the 24th International Conference on World
  Wide Web}. International World Wide Web Conferences Steering Committee, 2015,
  pp. 1067--1077.

\bibitem{kipf2016semi}
Thomas~N Kipf and Max Welling,
\newblock ``Semi-supervised classification with graph convolutional networks,''
\newblock {\em arXiv preprint arXiv:1609.02907}, 2016.

\bibitem{scarselli2009graph}
Franco Scarselli, Marco Gori, Ah~Chung Tsoi, Markus Hagenbuchner, and Gabriele
  Monfardini,
\newblock ``The graph neural network model,''
\newblock {\em IEEE Transactions on Neural Networks}, vol. 20, no. 1, pp.
  61--80, 2009.

\bibitem{chen2018fastgcn}
Jie Chen, Tengfei Ma, and Cao Xiao,
\newblock ``Fastgcn: fast learning with graph convolutional networks via
  importance sampling,''
\newblock {\em arXiv preprint arXiv:1801.10247}, 2018.

\bibitem{hamilton2017inductive}
Will Hamilton, Zhitao Ying, and Jure Leskovec,
\newblock ``Inductive representation learning on large graphs,''
\newblock in {\em Advances in Neural Information Processing Systems}, 2017, pp.
  1024--1034.

\bibitem{kingma2013auto}
Diederik~P Kingma and Max Welling,
\newblock ``Auto-encoding variational bayes,''
\newblock {\em arXiv preprint arXiv:1312.6114}, 2013.

\bibitem{rezende2014stochastic}
Danilo~Jimenez Rezende, Shakir Mohamed, and Daan Wierstra,
\newblock ``Stochastic backpropagation and approximate inference in deep
  generative models,''
\newblock {\em arXiv preprint arXiv:1401.4082}, 2014.

\bibitem{goodfellow2014generative}
Ian Goodfellow, Jean Pouget-Abadie, Mehdi Mirza, Bing Xu, David Warde-Farley,
  Sherjil Ozair, Aaron Courville, and Yoshua Bengio,
\newblock ``Generative adversarial nets,''
\newblock in {\em Advances in neural information processing systems}, 2014, pp.
  2672--2680.

\bibitem{you2018graphrnn}
Jiaxuan You, Rex Ying, Xiang Ren, William~L Hamilton, and Jure Leskovec,
\newblock ``Graphrnn: A deep generative model for graphs,''
\newblock {\em arXiv preprint arXiv:1802.08773}, 2018.

\bibitem{simonovsky2018graphvae}
Martin Simonovsky and Nikos Komodakis,
\newblock ``Graphvae: Towards generation of small graphs using variational
  autoencoders,''
\newblock {\em arXiv preprint arXiv:1802.03480}, 2018.

\bibitem{kipf2016variational}
Thomas~N Kipf and Max Welling,
\newblock ``Variational graph auto-encoders,''
\newblock {\em arXiv preprint arXiv:1611.07308}, 2016.

\bibitem{grover2018graphite}
Aditya Grover, Aaron Zweig, and Stefano Ermon,
\newblock ``Graphite: Iterative generative modeling of graphs,''
\newblock {\em arXiv preprint arXiv:1803.10459}, 2018.

\bibitem{lika2014facing}
Blerina Lika, Kostas Kolomvatsos, and Stathes Hadjiefthymiades,
\newblock ``Facing the cold start problem in recommender systems,''
\newblock {\em Expert Systems with Applications}, vol. 41, no. 4, pp.
  2065--2073, 2014.

\bibitem{chen2016variational}
Xi~Chen, Diederik~P Kingma, Tim Salimans, Yan Duan, Prafulla Dhariwal, John
  Schulman, Ilya Sutskever, and Pieter Abbeel,
\newblock ``Variational lossy autoencoder,''
\newblock {\em arXiv preprint arXiv:1611.02731}, 2016.

\bibitem{higgins2017beta}
Irina Higgins, Loic Matthey, Arka Pal, Christopher Burgess, Xavier Glorot,
  Matthew Botvinick, Shakir Mohamed, and Alexander Lerchner,
\newblock ``beta-vae: Learning basic visual concepts with a constrained
  variational framework,''
\newblock in {\em International Conference on Learning Representations}, 2017.

\bibitem{sen2008collective}
Prithviraj Sen, Galileo Namata, Mustafa Bilgic, Lise Getoor, Brian Galligher,
  and Tina Eliassi-Rad,
\newblock ``Collective classification in network data,''
\newblock {\em AI magazine}, vol. 29, no. 3, pp. 93, 2008.

\end{thebibliography}

\end{document}